\documentclass[runningheads]{llncs}

\usepackage[T1]{fontenc}
\usepackage{graphicx,verbatim}
\usepackage{amsmath,amssymb}
\usepackage{marvosym}

\begin{document}

\title{Clinically Aligned Geometry Constraints for Robust IVUS Vessel Boundary Segmentation}

\author{Yunshu Chen\inst{1,2} \and Litao Yang\inst{1,2}(\Letter) \and Giuseppe Di Giovanni\inst{3} \and Jordan Tan\inst{3} \and Deval Mehta\inst{1,2,4} \and Andrew Lin\inst{3} \and Derek Chew\inst{3} \and Masasi Fujino\inst{5} \and Julie Butters\inst{3} \and Stephen Nicholls\inst{3} \and Zongyuan Ge\inst{1,2} \and Kyung Hoon Cho\inst{3,6}(\Letter)}
 
\authorrunning{Chen, Yang, Di Giovanni et al.}
 
\institute{AIM For Health Lab, Monash University, Melbourne, Australia \and Department of Data Science and Artificial Intelligence, Faculty of IT, Monash University, Melbourne, Australia \and Monash University Victorian Heart Institute, Melbourne, Australia \and School of Computing Technologies, RMIT University, Melbourne, Australia \and National Cerebral and Cardiovascular Center, Suita, Osaka, Japan \and Department of Cardiology, Chonnam National University Hospital and Medical School, Gwangju, Republic of Korea \\
    litao.yang@monash.edu, kyung.cho@monash.edu}
 
\maketitle

\begin{abstract}
Intravascular ultrasound (IVUS) lumen and external elastic membrane (EEM)
segmentation is important for quantitative coronary plaque burden assessment.
Errors in lumen or EEM delineation directly propagate to plaque area, plaque
burden and geometric measurements. However, standard methods prioritising
overlap scores often suffer from boundary drift and topology errors, leading to
inaccurate clinical measurements. We present GeoCat, a geometry-consistent network that processes
5-frame IVUS clips using dual Cartesian--polar encoders with
cross-domain attention and temporal fusion. A differentiable
geometry consistency loss directly supervises clinically relevant
descriptors including diameters, orientations, and cross-sectional
areas. The model is trained on 12{,}242 annotated frames from 146
patients acquired with two commercial IVUS systems. We evaluate
performance using both segmentation accuracy and plaque-relevant clinical
metrics, including Dice/IoU, boundary measures (95HD (mm), ASSD), topology
violation rate, and clinical geometry errors ($d_{\max}$/$d_{\min}$, angles,
and areas). On our dataset, GeoCat achieves a Dice of 0.93, reduces 95HD to
0.14\,mm, and lowers topology violations to 1.0\%. Importantly, it
significantly improves geometric fidelity, yielding diameter errors of
0.13--0.16\,mm and angular errors of ${\sim}8$ degrees, supporting
reliable plaque burden quantification.

\keywords{IVUS segmentation \and temporal attention \and geometry consistency}
\end{abstract}

\section{Introduction}

Intravascular ultrasound (IVUS) is widely used for quantitative coronary plaque
burden assessment. Many clinical measurements, such as lumen area and external
elastic membrane (EEM) area, rely on accurate boundary delineation~\cite{ref_mintz2001}. Manual contouring across an entire IVUS pullback is time-consuming
and impractical in routine clinical workflows; consequently
clinicians typically analyse a limited number of selected frames.
This sparse analysis can overlook plaque burden changes and limits
comprehensive assessment along the vessel. Automated segmentation
addresses this bottleneck by reducing manual labour and minimising
inter-observer variability~\cite{ref_mintz2001,ref_balocco14}.
Beyond efficiency, accurate delineation enables quantitative
measurements of lumen and vessel diameter and cross-sectional area (CSA) to directly inform clinical decisions.
Percent atheroma volume (PAV) derived from these measurements are used to assess plaque regression or progression in therapeutic trials~\cite{ref_mintz2011}, while precise vessel
geometry guides stent sizing during percutaneous coronary
intervention~\cite{ref_yoon2012}. In typical datasets, labels are
sparse, but the underlying pullback sequence is dense, enabling
the use of nearby frames as context without additional annotation cost.

Recent studies have further explored annotation
efficiency~\cite{ref_autoannot25}, hierarchical feature
learning~\cite{ref_dah24}, and system-specific
adaptation~\cite{ref_kcj24}, but none jointly address temporal
consistency and differentiable geometry supervision. General
architectures such as U-Net~\cite{ref_unet} and
DeepLabV3+~\cite{ref_deeplabv3p} achieve reasonable overlap scores on IVUS
benchmarks~\cite{ref_dong23} but lack domain-specific
constraints~\cite{ref_coord21,ref_highres21}.
IVUS-tailored methods address different aspects of the problem.
Along the boundary refinement direction,
POLYCORE~\cite{ref_polycore24} represents contours as polygon
vertices and refines them with a graph network, reducing
Hausdorff distance but operating on single frames without
temporal consistency. Along the temporal modelling direction,
POST-IVUS~\cite{ref_postivus23} encodes multi-frame context
with a selective transformer and achieves strong Dice, yet its
loss function remains pixel-wise and does not explicitly
constrain geometric quantities. Along the geometry-aware
direction, Geo-UNet~\cite{ref_geounet24} introduces geometric
penalties in Cartesian space; however, extracting diameters and
orientations from Cartesian masks requires non-differentiable
operations such as connected-component analysis, limiting
end-to-end gradient flow. No existing method jointly addresses
temporal stability, dual-domain geometric reasoning, and
differentiable clinical metric supervision.

These gaps have practical consequences. Frame-wise methods
produce temporally inconsistent contours, complicating
longitudinal plaque tracking. Pixel-level losses do not
penalise clinically relevant quantities, a contour with
0.92 Dice can still yield diameter errors exceeding
0.3\,mm as we observe in our experiments. Polar reparameterisation
naturally makes diameter extraction differentiable, yet this
has not been exploited for end-to-end IVUS loss supervision.

We propose GeoCat to address these gaps. (1) To exploit temporal
continuity in IVUS pullbacks, we introduce a centre-query temporal
attention module that lets the target frame selectively attend to
neighbouring frames. (2) To combine the structured geometry of
polar coordinates with the spatial fidelity of Cartesian images,
we fuse features from dual-domain encoders via bidirectional
cross-attention. (3) To directly align segmentation with clinical
measurements, we propose a differentiable geometry consistency
loss that supervises diameters, orientations, and plaque-related
areas in the polar domain. On 12{,}242 frames from 146
patients, GeoCat reduces boundary error by 15\% and diameter MAE
by 12\% relative to the strongest prior method while halving
topology violations.

\section{Method}
\begin{figure}[t!]
    \centering
    \includegraphics[width=\textwidth]{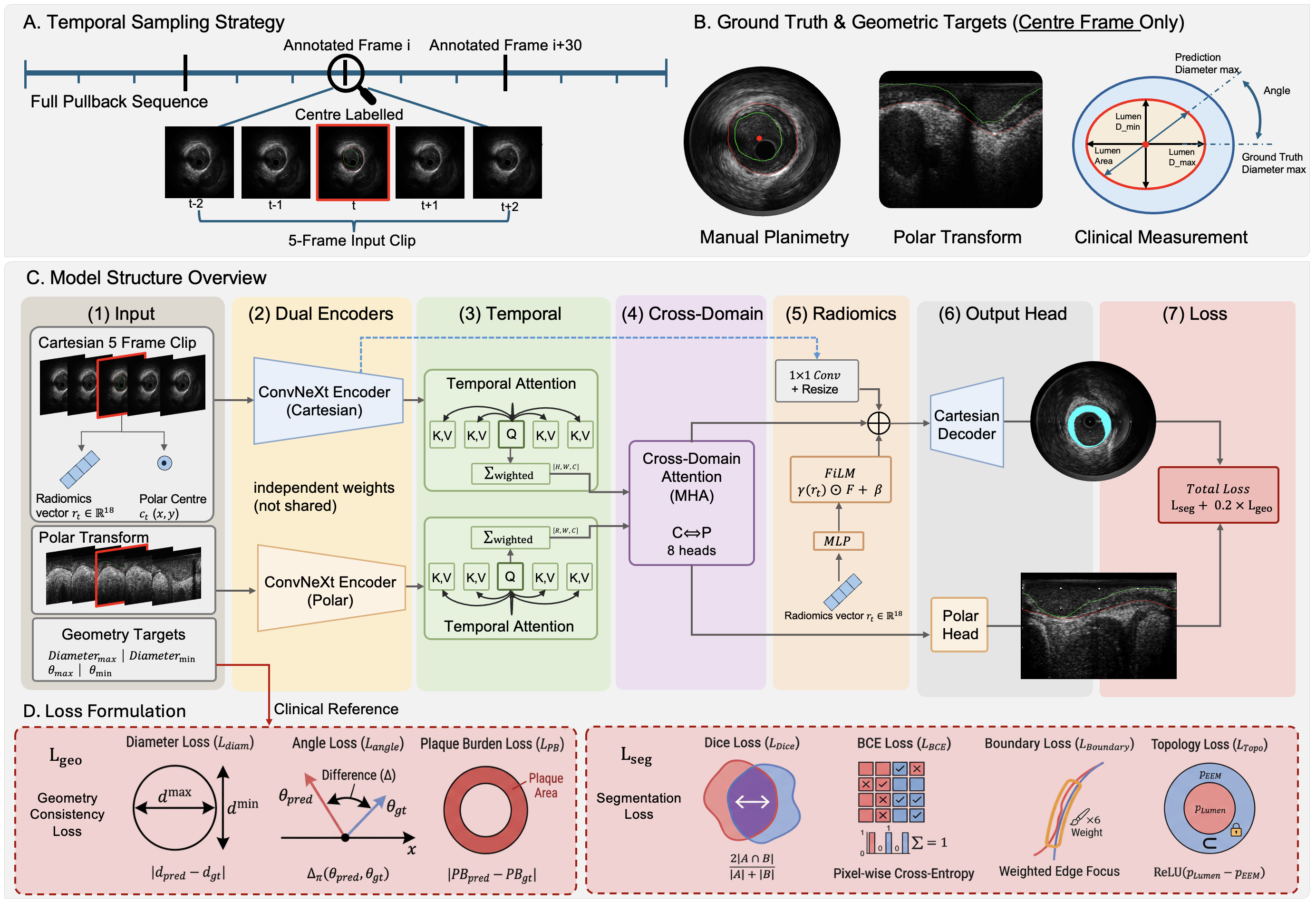}
\caption{GeoCat pipeline. (A)~Temporal clip sampling.
(B)~Polar reparameterisation and geometry target derivation.
(C)~Dual-encoder architecture with temporal attention,
cross-domain fusion, and FiLM conditioning.
(D)~Loss formulation.}
    \label{fig:arch}
\end{figure}
Fig.~\ref{fig:arch} illustrates our pipeline. GeoCat uses a 5-frame grayscale clip centred on the target frame as an input (Panel~A), plus an 18-dimensional radiomics descriptor. Two independent ConvNeXt-Base encoders
process each frame in parallel, one in Cartesian space and one in polar space;
their features are aggregated across time via a centre-query temporal attention
module and then aligned via bidirectional cross-domain attention. The Cartesian
decoder, conditioned on radiomics via FiLM, produces the final lumen and EEM
segmentation maps, while an auxiliary polar head trained jointly provides
supervision for the geometry consistency loss. 

\subsection{Preprocessing}
\paragraph{Polar transform.}
\label{sec:polar}
Each Cartesian frame is reparameterised to a polar grid
($A{=}512$ angle bins, $R{=}256$ radial steps, pixel spacing
$s{=}0.02$\,mm) centred at the image centre $c=(H/2, W/2)$ via differentiable
bilinear interpolation. Both domains are processed by
independent ImageNet-pretrained ConvNeXt-Base encoders.
\paragraph{Radiomics conditioning.}
Radiomics descriptors are extracted from the raw B-mode frame
without any segmentation mask. We select 18 features covering three groups of tissue appearance: 5 first-order intensity statistics capturing echogenicity, 6 GLCM
texture descriptors (averaged over 4 orientations) encoding speckle patterns, and 7 shape descriptors of the largest connected component above median intensity (area, perimeter, eccentricity, solidity, extent, major axis length, minor axis length). All features are z-normalised per fold. This feature set follows established radiomics taxonomies~\cite{ref_zwanenburg2020} and is kept compact to avoid overfitting while providing sufficient signal for cross-system adaptation via FiLM~\cite{ref_film}.

\subsection{Dual-Domain Temporal Architecture}
\paragraph{Temporal fusion.}
For $T = 5$ input frames, both encoders produce per-frame feature tensors at
Stage~2 ($C = 512$ channels, stride~16). Temporal aggregation uses three
components in sequence: a learnable temporal position embedding of shape
$[1, T, C, 1, 1]$; a depthwise-separable temporal convolution (kernel size~3,
residual connection); and a centre-query attention mechanism (Fig.~\ref{fig:arch}C,block~3) that computes a $T{\times}T$ 
affinity matrix via a compact MLP, extracts the centre-frame row, applies 
softmax to obtain per-frame weights, and produces a single spatial feature 
map $[B, C, H, W]$ as their weighted sum.

\paragraph{Cross-domain fusion.}
The temporally aggregated Cartesian and polar features are fused via
bidirectional multi-head cross-attention~\cite{ref_vaswani17} (Fig.~\ref{fig:arch}C,
8~heads, FFN expansion $4\times$, dropout~0.1). Spatial tokens
from each domain serve alternately as queries and keys/values
for the other, producing fused feature maps for each domain.
Residual connections preserve domain-specific information,
allowing each branch to retain its representation
while incorporating cross-domain cues.

\paragraph{Decoder.}
The fused Stage~2 Cartesian features are projected via a $1{\times}1$ 
convolution and residually added (weight $0.3$) to the centre-frame 
Stage~3 features, then modulated by FiLM~\cite{ref_film}(Fig.~\ref{fig:arch}C, block~5): an MLP maps $\mathbf{r}$ to channel-wise scale and shift parameters $(\gamma, \beta \in \mathbb{R}^{1024})$, applied as
$\gamma \odot \text{features} + \beta$. A 4-layer transposed-convolution decoder
upsamples to the input resolution, producing logits for lumen and EEM in
Cartesian space. In parallel, a 3-layer convolutional polar head produces logits
on the polar grid $[B, 2, 512, 256]$, used only for the geometry loss during
training.

\subsection{Differentiable Geometry Consistency Loss}
Pixel-wise overlap losses do not directly supervise clinically relevant 
geometric quantities such as maximum/minimum diameter, orientation, and 
cross-sectional area. We introduce a differentiable geometry consistency 
loss(Fig.~\ref{fig:arch}D, left) applied identically to two polar representations: the auxiliary polar 
head output, and the main Cartesian prediction reparameterised to polar 
coordinates via differentiable bilinear grid sampling; losses from both 
paths are averaged before back-propagation.

\paragraph{Soft boundary extraction.}
Let $p(\theta, r)$ denote the sigmoid probability at angle bin $\theta$ and
radius bin $r$ in the polar head output. For each angle $\theta$, the boundary
radius is extracted as a soft-argmin at the 0.5 probability crossing:
\begin{equation}
    r(\theta)
    = \sum_{r} \frac{\exp\!\left(-\lvert p(\theta,r)-0.5\rvert / \tau\right)}
                    {\sum_{r'} \exp\!\left(-\lvert p(\theta,r')-0.5\rvert / \tau\right)}
      \cdot r,
    \quad \tau = 0.05,
\end{equation}
yielding a differentiable boundary function
$r(\theta) \in \mathbb{R}^{A}$. The low temperature
($\tau{=}0.05$) concentrates weight on the bin nearest to the
0.5 crossing, bounding the approximation error to below one
radial bin ($0.02$\,mm).

The opposite diameter at angle $\theta$ is the full chord through the polar
centre:
\begin{equation}
    d(\theta) = r(\theta) + r\!\bigl((\theta+\pi)\bmod 2\pi\bigr).
\end{equation}
\paragraph{Soft extrema.}
Maximum and minimum diameters and their orientations are extracted via
temperature-scaled softmax weights ($\tau_{\text{topk}} = 0.1$). Soft extrema are computed with log-sum-exp stabilisation.
Let $w^{+}(\theta) = \exp(d(\theta)/\tau_{\text{topk}}) / Z^{+}$; then
\begin{equation}
    d^{\max} = \sum_{\theta} w^{+}(\theta)\, d(\theta),
    \qquad
    \theta^{\max} = \mathrm{atan2}\!\Bigl(
        \textstyle\sum_{\theta} w^{+}(\theta)\,\sin\theta,\;
        \sum_{\theta} w^{+}(\theta)\,\cos\theta
    \Bigr),
\end{equation}
and symmetrically for $d^{\min}$ and $\theta^{\min}$ using
$w^{-}(\theta) = \exp(-d(\theta)/\tau_{\text{topk}}) / Z^{-}$.

\paragraph{Cross-sectional area and plaque burden.}
The polar contour area is integrated as
\begin{equation}
    \mathrm{CSA} = \frac{1}{2} \sum_{\theta} r(\theta)^{2} \cdot \Delta\theta \cdot s^{2},
\end{equation}
where $\Delta\theta = 2\pi/A$. Plaque burden is defined as
\begin{equation}
    \mathrm{PB} = \frac{\mathrm{CSA}_{\mathrm{EEM}} - \mathrm{CSA}_{\mathrm{Lumen}}}
                       {\mathrm{CSA}_{\mathrm{EEM}}}.
\end{equation}

\paragraph{Geometry loss.}
$L_{\text{diam}}$ and $L_{\text{angle}}$ are computed for lumen
and EEM independently and averaged. $L_{\text{PB}}$ is computed
once from both structures jointly:
\begin{align}
L_{\text{diam}} &= \tfrac{1}{2}\!\left(L_{\text{diam}}^{\text{Lum}} + L_{\text{diam}}^{\text{EEM}}\right),\qquad
L_{\text{angle}} = \tfrac{1}{2}\!\left(L_{\text{angle}}^{\text{Lum}} + L_{\text{angle}}^{\text{EEM}}\right), \\
L_{\text{geo}} &= \lambda_d L_{\text{diam}} + \lambda_\theta L_{\text{angle}} + \lambda_{\mathrm{PB}} L_{\mathrm{PB}},
\end{align}
where, for structure $s \in \{\text{Lum}, \text{EEM}\}$,
\begin{equation}
    L_{\text{diam}}^{s} =
        \left| d^{\max,s}_{\text{pred}} - d^{\max,s}_{\text{gt}} \right|
      + \left| d^{\min,s}_{\text{pred}} - d^{\min,s}_{\text{gt}} \right|,
\end{equation}
\begin{equation}
    L_{\text{angle}}^{s} =
        \Delta_{\pi}\!\left(\theta^{\max,s}_{\text{pred}},\,\theta^{\max,s}_{\text{gt}}\right)
      + \Delta_{\pi}\!\left(\theta^{\min,s}_{\text{pred}},\,\theta^{\min,s}_{\text{gt}}\right),
\end{equation}
and $\Delta_{\pi}(a,b) = \min(\lvert a-b\rvert,\, \pi - \lvert a-b\rvert)$ is
the $\pi$-periodic circular distance; its subgradient at the
non-smooth point $\lvert a - b\rvert = \pi/2$ is never reached in
practice as both $\theta^{s}_{\text{pred}}$ and
$\theta^{s}_{\text{gt}}$ are continuous-valued. All diameter terms are in mm. Samples with
invalid or zero GT geometry are excluded via a validity mask. The weights are
$\lambda_{d} = 0.25$, $\lambda_{\theta} = 0.1$, and
$\lambda_{\mathrm{PB}} = 1.0$ (higher because PB is a ratio
in $[0,1]$ whereas diameters are in mm).

The total training loss is
$L = L_{\text{seg}} + 0.2\,L_{\text{geo}}$, where
$L_{\text{seg}}$ combines Dice~\cite{ref_milletari16},
BCE ($0.3$), boundary-weighted BCE ($0.2$), and topology penalty ($0.05$) that penalises
$\sum \mathrm{ReLU}(p_{\text{Lum}}{-}p_{\text{EEM}})$.
\section{Experiments}

\subsection{Dataset}
\label{sec:dataset}
The dataset is a private IVUS cohort comprising 12{,}242 annotated
cross-sections from coronary pullbacks of 146 patients acquired with two IVUS
systems (OptiCross 40\,MHz; Revolution 45\,MHz). Lumen/EEM planimetry was
performed at a fixed interval of 30 frames per pullback; only annotated frames
were used for supervision and evaluation. Analysts calibrated pixel spacing
using scanner-encoded 1\,mm grid marks and all contours underwent an
independent second-reader quality check. Frames where
side branches (2{,}994), severe calcification (1{,}931),
or major artefacts (2{,}604) prevented reliable planimetry
were not annotated; moderate cases are retained and flagged.
The remaining 12{,}242 frames constitute the benchmark.
Clinical geometry targets ($d_{\max}$/$d_{\min}$,
$\theta_{\max}$/$\theta_{\min}$, CSA, PB) were derived from
contours using standard definitions~\cite{ref_mintz2001}
and computed in the polar domain with respect to the image
centre $c=(H/2,W/2)$.

\subsection{Implementation details}
\label{sec:setup}
We hold out 10\% of patients as a test set; the remaining
patients are split into 5 folds for cross-validation. Same-patient scans are always co-assigned to prevent leakage; all methods share this protocol. We trained POLYCORE using its official code and re-implemented
POST-IVUS and Geo-UNet following their papers, all under our
protocol for fair comparison. All methods use the same two-class, identical $L_{\text{seg}}$, augmentation, resolution
($512 \times 512$), and cross-validation splits. POST-IVUS and
GeoCat receive 5-frame clips; all other methods receive single
frames. Post-processing (largest connected component, hole
filling) is applied uniformly. ConvNeXt (1f)~\cite{ref_convnext}
is a single-frame baseline using the same backbone as GeoCat,
included to isolate temporal, polar, and geometry contributions.
Each annotated frame is the centre of a 5-frame input clip drawn
from the dense pullback; supervision applies only to the centre
frame.

Both ConvNeXt-Base encoders are initialised from ImageNet-1K
weights, with a learnable $\mathrm{Conv2d}(1{\to}3,\,k{=}1)$
adapter for grayscale input. Training uses AdamW
(lr $5 \times 10^{-5}$, weight decay $10^{-5}$), cosine annealing
with a 5-epoch linear warmup, batch size~4, for 80~epochs.
Augmentation is applied consistently across all 5~frames:
horizontal/vertical flipping ($p=0.5$), rotation
$[-15^{\circ}, +15^{\circ}]$, zoom $[0.9, 1.1]$, and Gaussian
noise ($\sigma=0.05$, $p=0.5$). Geometry ground truth is
recomputed after spatial transforms.
For segmentation quality we report Dice, IoU, 95HD (mm),
ASSD (mm), and topology violation rate. For clinical geometry
we report MAE of $d_{\max}$, $d_{\min}$ (mm), angular MAE of
$\theta_{\max}$, $\theta_{\min}$ (degrees), and CSA MAE for
EEM and lumen (mm$^{2}$). All evaluation is in the Cartesian
domain.

\begin{table}[t]
\caption{Segmentation results (mean $\pm$ std, 5 folds),
macro-averaged over lumen and EEM. Viol.: lumen not
contained in EEM. \dag~temporal context.}
\label{tab:comparison}
\centering
\setlength{\tabcolsep}{3.0pt}
{\fontsize{8}{9}\selectfont
\begin{tabular}{lccccc}
\hline
Method & Dice$\uparrow$(\%) & IoU$\uparrow$ & 95HD$\downarrow$ (mm) &
ASSD$\downarrow$ (mm) & Viol.$\downarrow$ (\%) \\
\hline
U-Net~\cite{ref_unet}              & 83.5$\pm$2.1 & 0.717$\pm$0.028 & 0.304$\pm$0.112 & 0.102$\pm$0.015 & 5.72$\pm$1.84 \\
UNet++~\cite{ref_unetpp}           & 84.6$\pm$1.9 & 0.733$\pm$0.024 & 0.285$\pm$0.095 & 0.095$\pm$0.012 & 4.92$\pm$1.65 \\
DeepLabV3+~\cite{ref_deeplabv3p}   & 87.4$\pm$1.5 & 0.776$\pm$0.019 & 0.246$\pm$0.103 & 0.085$\pm$0.009 & 4.15$\pm$1.20 \\
ConvNeXt (1f)~\cite{ref_convnext}                     & 88.9$\pm$1.1 & 0.800$\pm$0.014 & 0.205$\pm$0.068 & 0.072$\pm$0.007 & 3.29$\pm$0.95 \\
\hline
Geo-UNet~\cite{ref_geounet24}      & 89.5$\pm$1.3 & 0.823$\pm$0.016 & 0.216$\pm$0.072 & 0.076$\pm$0.008 & 3.82$\pm$1.05 \\
POLYCORE~\cite{ref_polycore24}     & 90.4$\pm$0.9 & 0.832$\pm$0.011 & 0.195$\pm$0.055 & 0.065$\pm$0.006 & 2.56$\pm$0.82 \\
POST-IVUS\dag~\cite{ref_postivus23}& 91.9$\pm$0.8 & 0.849$\pm$0.010 & 0.168$\pm$0.042 & 0.051$\pm$0.007 & 1.97$\pm$0.55 \\
\hline
\textbf{GeoCat (Ours)}\dag         & \textbf{93.4$\pm$0.6} & \textbf{0.875$\pm$0.008} & \textbf{0.143$\pm$0.035} & \textbf{0.049$\pm$0.005} & \textbf{1.00$\pm$0.28} \\
\hline
\end{tabular}}
\end{table}

\begin{table}[t]
\caption{Clinical geometry results (mean $\pm$ std, 5 folds).
Diameters and angles macro-averaged over lumen/EEM;
CSA reported per structure.}
\label{tab:clinical}
\centering
\setlength{\tabcolsep}{4.5pt}
{\fontsize{8}{9.5}\selectfont
\begin{tabular}{lccccc}
\hline
Method & $d_{\max}$ MAE & $d_{\min}$ MAE & $A_{\text{EEM}}$ MAE & $A_{\text{Lum}}$ MAE & $\Delta\theta$ MAE \\
 & (mm) & (mm) & (mm$^2$) & (mm$^2$) & (deg) \\
\hline
U-Net~\cite{ref_unet}              & 0.300$\pm$0.082 & 0.260$\pm$0.075 & 6.20$\pm$1.85 & 4.70$\pm$1.42 & 18.0$\pm$4.6 \\
UNet++~\cite{ref_unetpp}           & 0.280$\pm$0.076 & 0.240$\pm$0.068 & 5.80$\pm$1.60 & 4.40$\pm$1.25 & 16.9$\pm$4.2 \\
DeepLabV3+~\cite{ref_deeplabv3p}   & 0.250$\pm$0.065 & 0.215$\pm$0.059 & 4.90$\pm$1.35 & 3.70$\pm$1.10 & 15.4$\pm$3.7 \\
ConvNeXt (1f)~\cite{ref_convnext}                       & 0.230$\pm$0.058 & 0.195$\pm$0.052 & 4.50$\pm$1.22 & 3.30$\pm$0.95 & 14.2$\pm$3.3 \\
\hline
Geo-UNet~\cite{ref_geounet24}      & 0.220$\pm$0.061 & 0.185$\pm$0.048 & 4.20$\pm$1.15 & 3.40$\pm$0.88 & 14.6$\pm$3.5 \\
POLYCORE~\cite{ref_polycore24}      & 0.198$\pm$0.045 & 0.158$\pm$0.042 & 3.45$\pm$0.92 & 2.90$\pm$0.75 & 12.8$\pm$2.8 \\
POST-IVUS$^\dag$~\cite{ref_postivus23} & 0.182$\pm$0.038 & 0.162$\pm$0.035 & 2.95$\pm$0.85 & 2.35$\pm$0.62 & 11.2$\pm$2.4 \\
\hline
\textbf{GeoCat (Ours)$^\dag$}      & \textbf{0.160$\pm$0.025} & \textbf{0.132$\pm$0.022} & \textbf{2.65$\pm$0.68} & \textbf{1.95$\pm$0.54} & \textbf{8.6$\pm$1.8} \\
\hline
\end{tabular}}
\end{table}

\subsection{Segmentation and Clinical Geometry Results}

Tables~\ref{tab:comparison} and~\ref{tab:clinical} summarise all results. DeepLabV3+ is the best general method (Dice 0.874) but
topology violations reach 4.1\%. Among IVUS-specific methods,
POST-IVUS achieves the highest prior Dice (0.919), yet all
baselines exceed 1.9\% violations as none combine temporal
consistency with geometric constraints.

Compared to Geo-UNet, which also applies geometry constraints
but operates in Cartesian space on single frames, GeoCat reduces
$d_{\max}$ MAE from 0.220 to 0.160\,mm and topology violations
from 3.8\% to 1.0\%, confirming that polar-domain geometry
supervision and temporal context each contribute independently.
Against POST-IVUS (the strongest prior method), GeoCat improves
Dice by 0.015 and 95HD by 0.025\,mm. 

\begin{figure}[t!]
    \centering
    \includegraphics[width=0.97\textwidth, height=5.5cm, keepaspectratio]{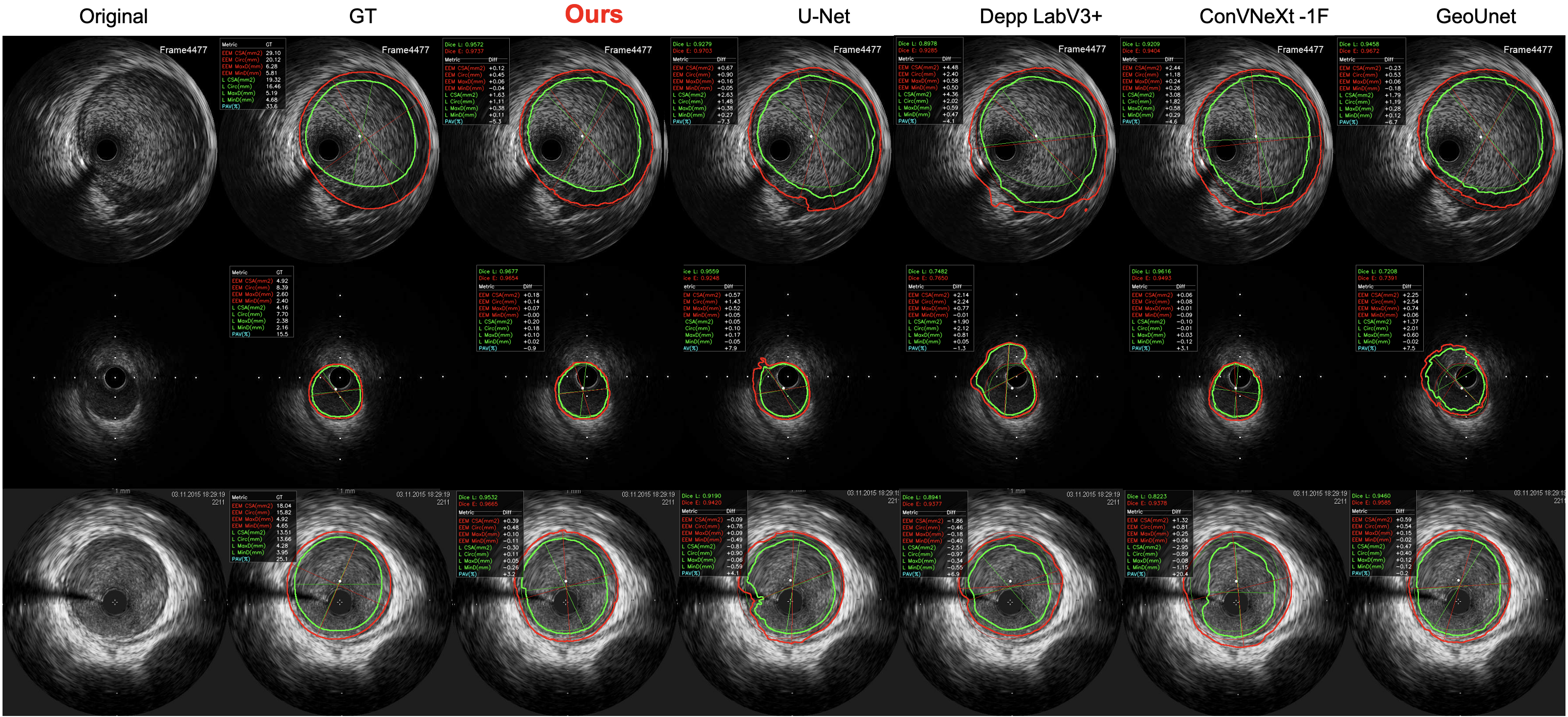}
    \caption{Qualitative comparison on three test cases. EEM/lumen
contours in red/green; inset tables show per-method clinical
errors. GeoCat yields smoother boundaries and lower diameter
errors, especially in rows~2--3.}
    \label{fig:qualitative}
\end{figure}

Fig.~\ref{fig:qualitative} shows three representative cases.
In row~1 (normal plaque), all methods produce similar contours,
but GeoCat yields a smoother EEM boundary. In row~2 (acoustic
shadow from calcification), single-frame methods underestimate
the EEM extent behind the shadow, while GeoCat recovers the
boundary via temporal context. In row~3 (large eccentric plaque),
baseline methods produce lumen contours that leak into the plaque
region, whereas GeoCat maintains topological consistency with
a $d_{\max}$ error of only 0.08\,mm. 

\begin{table}[t]
\caption{Cumulative ablation on the test set (mean $\pm$ std across 5 folds).A: ConvNeXt-1f; B: +5f-stack; C: +centre-attn; D: +polar fusion; E: +$L_{\text{geo}}$; F: GeoCat.}
\label{tab:ablation}
\centering
\setlength{\tabcolsep}{2.2pt}
\renewcommand{\arraystretch}{0.95}
{\fontsize{8}{9}\selectfont
{%
\begin{tabular}{lccccccccc}
\hline
Config & 5f & Ctr. & Polar & $L_{\text{geo}}$ & FiLM &
Dice$\uparrow$ & 95HD$\downarrow$ & Viol.$\downarrow$ & $d_{\max}\downarrow$ \\
& stack & attn & branch & & & (\%)& (mm) & (\%) & MAE (mm) \\
\hline
A  & & & & & & 88.9$\pm$1.5 & 0.205$\pm$0.085 & 3.29$\pm$1.15 & 0.230$\pm$0.075 \\
B    &\checkmark& & & & & 90.6$\pm$1.2 & 0.185$\pm$0.072 & 2.60$\pm$0.95 & 0.215$\pm$0.062 \\
C &\checkmark&\checkmark& & & & 91.4$\pm$1.0 & 0.172$\pm$0.055 & 2.10$\pm$0.75 & 0.205$\pm$0.051 \\
D &\checkmark&\checkmark&\checkmark& & & 92.6$\pm$0.8 & 0.152$\pm$0.045 & 1.25$\pm$0.42 & 0.188$\pm$0.040 \\
E &\checkmark&\checkmark&\checkmark&\checkmark& & 92.9$\pm$0.7 & 0.148$\pm$0.038 & 1.10$\pm$0.35 & 0.172$\pm$0.032 \\
F(ours)        &\checkmark&\checkmark&\checkmark&\checkmark&\checkmark&\textbf{93.4$\pm$0.6}&\textbf{0.145$\pm$0.038}&\textbf{1.00$\pm$0.28}&\textbf{0.160$\pm$0.025}\\
\hline
\end{tabular}
}}
\end{table}


\subsection{Ablation Study}

Table~\ref{tab:ablation} isolates each component. Temporal context
(B, C) consistently reduces boundary error, with centre-query
attention (C) outperforming channel stacking (B) by selectively
weighting informative neighbours. Adding polar fusion (D) reduces
topology violations from 2.1\% to 1.25\% before any explicit
geometry supervision, indicating that polar features impose a soft radial ordering that discourages lumen-outside-EEM configurations. The geometry
consistency loss (E) then provides explicit supervision:
$d_{\max}$ MAE drops from 0.188 to 0.172\,mm, confirming that pixel-wise losses alone do not constrain diameter orientation. FiLM conditioning (F) yields further uniform gains. Notably, $L_{\text{geo}}$ alone (D$\to$E) reduces $d_{\max}$ MAE by 8.5\% while Dice improves only 0.3\%, confirming that overlap metrics underestimate clinical boundary errors.

\section{Conclusion}

GeoCat combines dual Cartesian--polar encoders, temporal
attention, and a differentiable geometry loss to jointly improve
segmentation overlap and clinical measurements for IVUS lumen/EEM
delineation, reducing topology violations to 1.0\% and diameter errors 
to 0.13--0.16\,mm, which are within clinically 
actionable margins for stent sizing~\cite{ref_yoon2012}. Limitations include the catheter-centring assumption and ${\sim}10\times$ inference cost (142\,ms, A100); future work will explore lightweight backbones and predicted polar centres. As no public IVUS benchmark remains available, we will release code, annotations, and evaluation splits.


\end{document}